\pdfoutput=1

\documentclass[11pt]{article}

\usepackage{lrec-coling2024} 
\usepackage{graphicx}
\usepackage{subcaption}
\usepackage{times}
\usepackage{latexsym}
\usepackage{booktabs}
\usepackage[T1]{fontenc}

\usepackage[utf8]{inputenc}

\usepackage{microtype}

\title{Detecting Conceptual Abstraction in LLMs}

\name{Michaela Regneri, Alhassan Abdelhalim, Sören Laue } 
\address{
         Universität Hamburg \\ Dept. of Computer Science, Machine Learning Research Group\\ Hamburg, Germany\\
         \{michaela.regneri,alhassan.abdelhalim,soeren.laue\}@uni-hamburg.de}

\abstract{We present a novel approach to detecting noun abstraction within a large language model (LLM). Starting from a psychologically motivated set of noun pairs in taxonomic relationships, we instantiate surface patterns indicating hypernymy and analyze the attention matrices produced by BERT. We compare the results to two sets of counterfactuals and show that we can detect hypernymy in the abstraction mechanism, which cannot solely be related to the distributional similarity of noun pairs. Our findings are a first step towards the explainability of conceptual abstraction in LLMs.\\
\newline 
 \Keywords{transformers, abstraction, attention, explainability} }

\begin{document}
\maketitleabstract


\section{Introduction}
Large Language Models (LLMs) have emerged as a powerful tool for a plethora of applications. State-of-the-art LLMs are based on the transformer architecture~\cite{NIPS2017_3f5ee243} that can directly generate text sequences (like chatbots), translate texts, or lend their outcomes to other downstream tasks. Due to their versatile functionality, LLMs are often distributed as pre-trained black-box models, which can then be fine-tuned to specific needs.

While LLMs surpass the performance of simpler models, they are far less explainable due to the intransparent nature of their complex architecture. More explainability can be crucial in multiple applications, e.g., if models must adhere to some governance to prevent bias or build more data-efficient models. Especially in the context of trustworthy AI, one central open research question is how these models' excellent output is achieved and whether the mechanisms internally employed in LLMs reassemble those present in humans.

We provide an analysis examining whether simple linguistic abstraction mechanisms are present in a large language model.
For humans, relations like hypernymy (\emph{ravens} are \emph{birds}) are essential for linguistic understanding and generalization. LLMs also necessarily employ some kind of abstraction and generalization, but most likely not exactly in the same way as humans do. With our experiments, we add one more step toward representing hypernym relationships within large language models and, thus, their capacity to use human-like abstraction mechanisms for generalization.
Specifically, we test BERT \cite{devlin-etal-2019-bert} for its attention patterns related to taxonomic hypernyms and compare this to unrelated noun pairs with either high or low semantic similarity. We draw our test data from a psychologically motivated data set of human associations, which lends itself to examining hypernym pairs with high cognitive saliency. Our results show that BERT represents this kind of abstraction within its attention module.

Our main contribution consists of clear evidence that LLMs infer linguistic abstraction and that this inference goes beyond semantic similarity. For this, we provide both a method and a dataset to show the attention patterns of LLMs for semantic hypernymy and separate them from counterfactuals matched by semantic similarity and abstraction level.


\section{Background and Related Work}\label{background}
In the past years, the capabilities of LLMs have been enhanced tremendously. With transformer models~\cite{NIPS2017_3f5ee243} as an architectural basis, LLMs are trained on vast amounts of text and optimized to predict the next word in a sequence or the following sentence in a discourse. The resulting models have many applications and can model many linguistic phenomena known to be crucial for human language~(\citealp{doi:10.1073/pnas.1907367117}).
When analyzing the emergence of linguistic phenomena, a particular focus lies on the self-attention mechanism of transformers. Self-attention is a step in the encoder part of these language models. It maps the input sequence to a weighted representation of itself and thus, intuitively speaking, reveals the sequence's focal points relevant to generating its follow-up. Self-attention consists of multiple so-called heads, which act in parallel on the sequence and are multiplied in several layers (see~\citet{NIPS2017_3f5ee243} for details). The grid of attention heads with the individual scores they attach to the sequence is often treated as a proxy for the information encoded in the transformer. For some discussion on how far this is possible, see, e.g., \citet{jain-wallace-2019-attention} and \citet{wiegreffe-pinter-2019-attention}.
There are two types of approaches that recover linguistic structure in LLMs: One performs end-to-end evaluation by disabling or manipulating single attention heads and evaluating the performance change for different tasks \cite[e.g.]{kovaleva-etal-2019-revealing}. Others look directly into the attention patterns, which we also do.
    \begin{table}
        \centering
        \footnotesize
        \begin{tabular}{l p{65mm}}
        \toprule
        \#No & Pattern \\ \midrule
          1   & [hypo]s are [hyper]s.  \\
          2   & That [hypo] is [a(n)] [hyper]. \\
          3 & I like [hypo]s and other [hyper]s. \\
          4& The [hypo], which was the largest [hyper] among them, stood out.\\
          5 &  I like [hypo]s, particularly because they are [hyper]s.\\ \bottomrule
        \end{tabular}
        \caption{Hypernymy patterns, with [hypo] and [hyper] as slots for target and the feature concepts respectively. Plurals are indicated with \emph{s} and [a(n)] is a determiner. }
        \label{tab:patterns}
    \end{table}
\citet{doi:10.1098/rstb.2019.0307} shows an overview of abstraction and compositionality in artificial neural networks. Many approaches use artificial languages and small models \cite[e.g.]{pmlr-v80-lake18a,ijcai2020p708}, others also test pre-trained LLMs like BERT~\cite{devlin-etal-2019-bert}. See \citet{sajjad-neuron-survey} for an overview. Several approaches have found evidence for linguistic knowledge within BERT. For instance, \citet{chen-etal-2023-causal} prompt the model with correct and counterfactual data and then infer BERT's abstraction capabilities.
Only a few approaches show results for deep semantic knowledge directly within the attention mechanism. \citet{dalvi2019neurox} try to discover latent concepts in BERT, which are essentially hypernyms and their derivable hyponyms. 
In our approach, we focus on hypernym-hyponym relations between nouns as one central linguistic abstraction phenomenon. For collecting hypernyms by prompting, \citet{hanna-marecek-2021-analyzing} present an experiment in which BERT outperforms other unsupervised algorithms in the collection of common hypernyms, which suggest that the model at least has the capacity to user hypernymy. This raises the questions on whether and how this is also internally represented in the trained model.
To the best of our knowledge, there is no approach that characterizes attention patterns for generic hypernyms, especially no approach that distinguishes taxonomic relationships from pure semantic similarity.
We take another step towards understanding conceptual abstraction in LLMs, evaluate the attention patterns related to true and counterfactual hypernyms, and show that the effects must be related to abstraction rather than similarity.

\section{Data}\label{data}

We create a data set of noun pairs that are in a hypernymy relationship, and two sets of counterfactual pairs (which are not hypernyms). In order to construct example sentences, we manually create patterns that typically express hypernymy in their surface form and instantiate them with the noun pairs.
\subsection{Positive Examples}
We extract hyponym-hypernym pairs from McRae's feature norms~\cite{featurenorms_2005}. The feature norms contain pairs of concepts (originally stimuli) and features (human associations), annotated with semantic relationships. The pre-selection gives us more salient pairs of terms than a full-fledged taxonomy and should also be recognizable as strongly related by a large language model.
For our data of valid noun pairs, we select all pairs of concepts and features labeled with a ``superordinate'' relationship in the feature norms. These concept-feature pairs have the target concept as a hyponym and the feature concept as a hypernym (e.g., \emph{raven} and \emph{bird}). The dataset can conatin multiple hypernyms for a concept with different levels of abstraction, e.g \emph{raven} and \emph{bird} as well as \emph{raven} and \emph{animal}. We include all such pairs in the dataset and balance them later with counterfactuals with similar degree of abstraction.

\subsection{Creating Counterfactuals}
We create two counterfactual sets of noun pairs, which are not in a hypernymy relationship and thus will produce invalid sentences within our patterns. Using WordNet~\cite{Fellbaum1998}, we generate the pairs by either sister terms of the feature concept from the positive examples (\emph{negative examples}), which are matched by the level of abstraction of the feature concepts, or terms which share a hypernym with the target concepts (\emph{sister terms}), which approximately match the level of similarity of the positive examples.
With those two sets, we want to exclude spurious effects from just measuring semantic similarity or differences in concept abstraction level (and thus indirectly also frequency).
\subsubsection*{Negative Examples}
For the first set of counterfactuals, we elicit noun pairs in which the feature concept is on the same level of generality as the hypernym in the first set. For instance, for the positive example \emph{raven -- animal}, we might choose \emph{raven -- person}. If there are multiple hypernyms for the same concept, we select an appropriate counterfactual for each individual hypernym. In detail, we proceed as follows:
\begin{enumerate}
    \item We map each positive example to the WordNet synsets by extracting those synset pairs that contain the respective lemmas and stand in a (direct or inherited) hypernymy relationship in WordNet.  
        \item For each feature synset, we select a sister term (a synset sharing a parent node), which is no hypernym of the target concept (in the example, we pick a sister term of the \emph{alligator} synset). To avoid effects from low-frequency words, we select the most frequent lemma from those sister synsets as a counterfactual (in the example \emph{person}).
\end{enumerate}
\subsubsection*{Sister Terms}
Our second counterfactual set controls for the level of semantic similarity within the positive examples. Hyponyms and their hypernyms are often distributionally very similar \cite{pado-lapata-2007-dependency}, especially salient ones. To measure whether we really find differences related to violations of taxonomic rules or just effects due to high semantic similarity, we pick a sister term in WordNet for each of the original target concepts (e.g., \emph{raven -- crow}, which are both hyponyms of \emph{bird}). As for the negative examples, we choose the most frequent sister term lemma.
Sister terms usually share many contexts, so we expect effects due to semantic similarity to be shared between the positive examples and the sister terms.
\subsection{Creating Test Sentences}
As input for the LLM, we create test sentences that express a taxonomic relationship directly or indirectly. First, we manually create a set of five sentence patterns that exhibit hypernymy relationships, partially inspired by the patterns used by~\citet{hearst-1992-automatic} to extract hyponym-hypernym pairs automatically from large text corpora. We vary the simplicity and saliency of the patterns to control for those effects. Table~\ref{tab:patterns}  shows the set of patterns.


We instantiate our patterns with the noun pairs from all three sets, resulting in 3425 examples per set. The results are sentences like \emph{I like ravens and other animals} (positive example), and \emph{I like ravens and other people} (negative example) and \emph{I like ravens and other crows} (sister term). 
We provide all data sets for reference. 


\begin{figure*}
    \centering
    \begin{subfigure}{0.29\textwidth}
        \includegraphics[width=\linewidth]{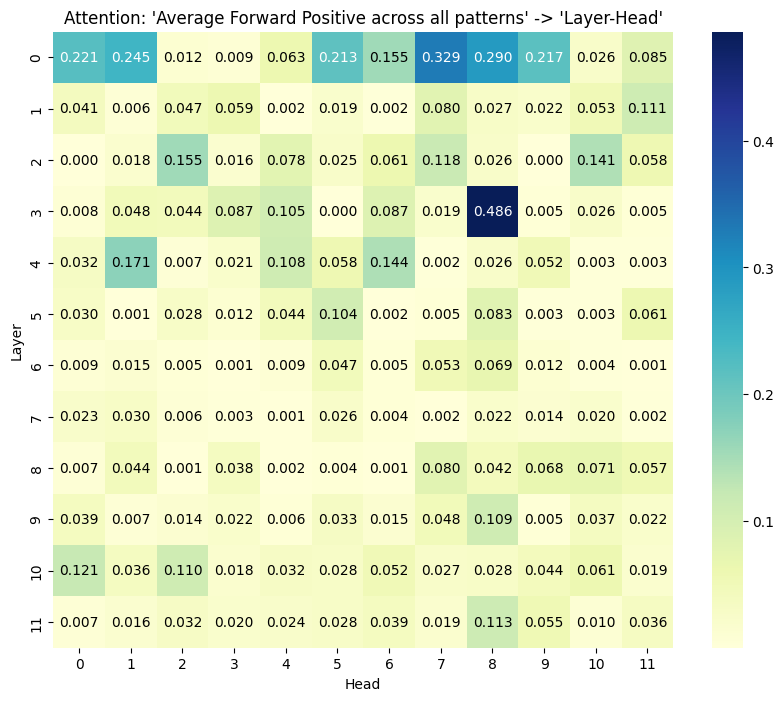}
        \caption{Forward positive}
    \end{subfigure}\qquad
    \begin{subfigure}{0.29\textwidth}
        \includegraphics[width=\linewidth]{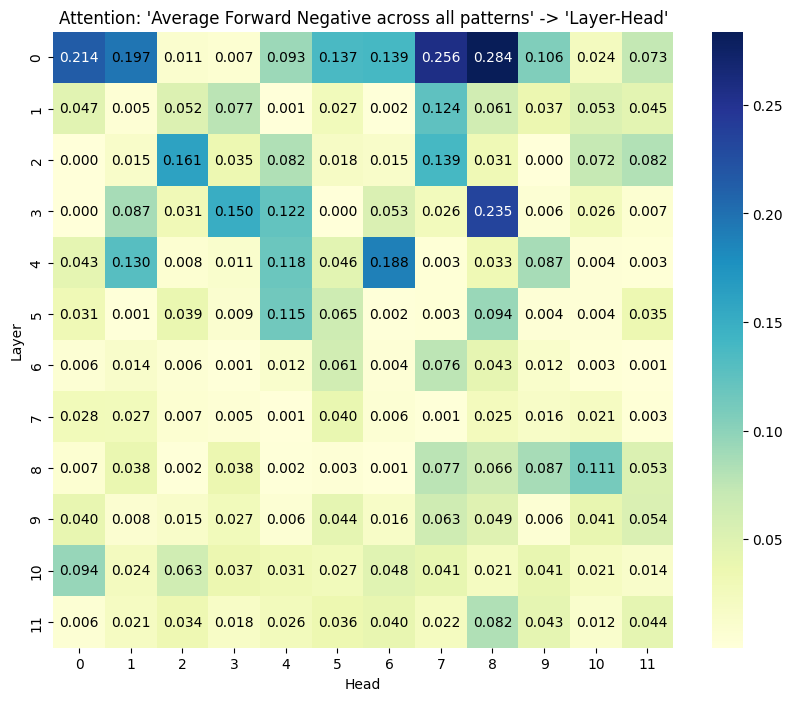}
        \caption{Forward negative}
    \end{subfigure}\qquad
    \begin{subfigure}{0.29\textwidth}
        \includegraphics[width=\linewidth]{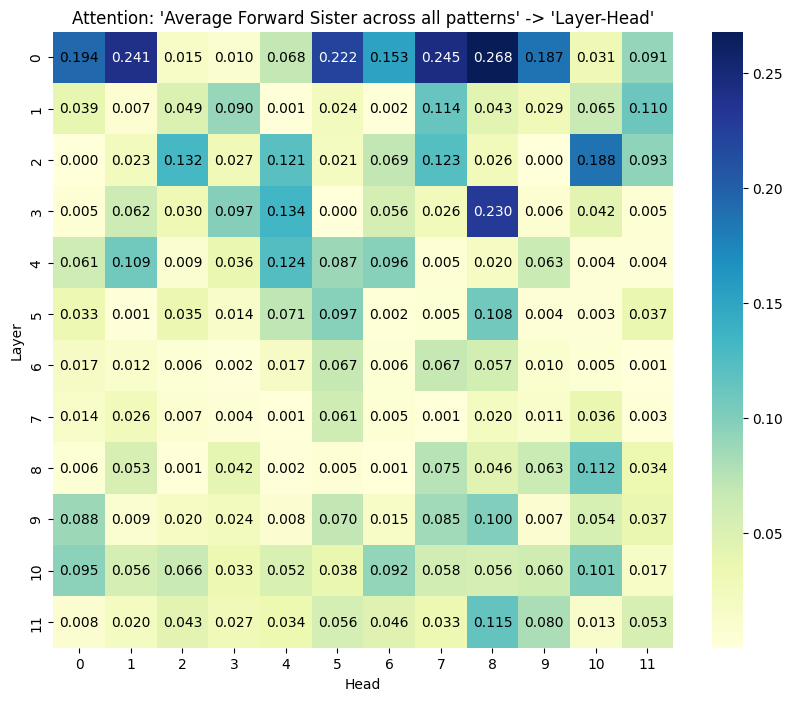}
        \caption{Forward sister}
    \end{subfigure}

    \caption{Attention maps for hyponyms and hypernyms averaged across all patterns.}\label{fig:averageattention}
\end{figure*}
\section{Hypernymy within BERT}\label{experiments}
We analyze whether or not hypernymy has a correlation with BERT's attention mechanism.  After visualizing the attention for all datasets, we separate them via a linear classifier. For all experiments, we use BERT-large in the monolingual English version. 

\subsection{Attention Matrices and Clustering}
For each sentence, we extract the self-attention values from BERT. We restrict our analysis to the forward-looking attention between our target and feature tokens. Each sentence is represented by a 12x12 attention matrix (with 12 layers of 12 attention heads per layer).
BERT's tokenizer breaks up some of our pluralized tokens, which makes the attention between a split token and a complete token incomparable. For the sake of simplicity, we discard all examples in which one of our tokens in focus is split up.

Figure~\ref{fig:averageattention} gives a high-level overview of the results for each data set. For this visualization, we average the attention between target and feature concepts (resp. their sequence position) over all examples. Each cell in a heatmap corresponds to one attention head (x-axis) in one layer (y-axis). Dark colors indicate a high average activation of the attention head.
Intuitively, we see that the three sets differ, so concept clashes in the negative set and the sister terms do expose different attention patterns than the salient hypernyms. Further, the overall attention seems lower in the positive setting than in the other two control settings. This suggests that higher attention here denotes some form of surprise for unexpected semantic constructions.

To validate how well the three sets are separable, we employ logistic regression to recover the three test sets automatically.
Each data point is the attention matrix of one example sentence, flattened into a 144-dimensional vector. For classification, we use the standard implementation of logistic regression from scikit-learn~\cite{sklearn} with all default parameter settings, setting the number of iterations to 1000 and the regularization parameter C to 1.
We also perform a pairwise comparison of the different sets to understand how similar levels of abstraction (sister terms vs. negative) or similar levels of semantic similarity (positive vs. sister terms) of the test tokens influence the separability of the examples.

\subsection{Results}
 We find a prediction accuracy of 0.75 on the test sets for our overall comparison and similar scores for the pairwise separation (Table~\ref{tab:resultsCluster}). Our three sets are equal-sized, so a random baseline would return an accuracy of about 0.33.
\begin{table}[httb]
    \centering
    \small
    \begin{tabular}{lc}
    \toprule
         Sets  & Acc.\\
         \midrule
         All three  &0.75 \\
         Pos. vs. Neg.&\textbf{0.88}\\
         Pos. vs. Sisters  &0.84 \\
         Neg. vs. Sisters  &0.85\\
         \bottomrule
    \end{tabular}
    \caption{Accuracy for predicting the test sets.}
    \label{tab:resultsCluster}
\end{table}
All scores indicate a substantial difference in attention patterns in the three sets. The positive and the negative examples are well separated. Here, we see the semantic type clash for non-hypernyms in a hypernymy pattern and the low semantic similarity of the target and feature concept. The sister terms are equally well distinguishable from both positive and negative examples, but the set is less well recoverable than the positive examples. This means that the differences we see between positive and negative examples must be due to something different than semantic similarity because the sister terms are distributionally very similar to their matched positive examples. Further, the attention seems to represent the subtle difference between the two sets of counterfactuals internally, which points to interesting research questions on the level of abstraction within the transformer models.

\subsection{Limitations}
Our approach takes a first step towards understanding linguistic abstraction in transformer models. Our experiments have several technical limitations and limitations in the interpretability of the results.

First, we restrict ourselves to taxonomic hypernymy of nouns, which is only a small part of abstraction. Within this theoretical limitation, our dataset is also limited to the hyponym-hypernym pairs from the feature norms we used as our source.

The restriction to a dictionary-based definition of abstraction also affects our dataset. When assembling the counterfactuals fitted to the input data, we found that some of our counterfactuals are strictly speaking no hypernyms, but colloquially still treated as such, e.g., \emph{spatula -- tool}, or \emph{barn -- shelter}. We leave those examples in the dataset, which might have influenced our results.

Further limitations of our input data result from our handling of tokenized words. We filter all words that are split up by the BERT tokenizer. There are several approaches that recombine subword tokens into whole words. Unfortunately, no standard approach fits all applications, so in future work, the most suitable way to retrieve whole words from subwords should be tested and applied.

Lastly, like for every probing approach, the interpretability of our results is debatable. We have shown that words in a hypernymy relationship give rise to attention matrices that are well distinguishable from counterfactuals, which are semantically wrong assertions. One can argue that the results on our dataset mainly show that we can distinguish salient sentences from absurd ones. We think that the least our results show is that there is something that is regularly attached to hypernymy that the transformer learned. Otherwise, we would not be able to separate the two sets of counterfactuals, which both consist of unlikely sentences and which are not distinguishable in their degree of oddity (\emph{I like ravens and other crows} is about as wrong as \emph{I like ravens and other people}). The only nuance between those counterfactuals is the degree of abstraction in the target words. Moreover, they both are well separable from the correct sentences when matched on the level of abstraction. So, while we cannot (and did not) claim that we found the attention pattern that completely explains how taxonomic abstraction works in transformers, we can claim that there is more than semantic similarity and reasonable content that makes the differences we measure.


\section{Summary and Future Work}\label{conclusion}
Our experiments show an initial indicator for linguistic, conceptual abstraction in the attention mechanism of LLMs. Based on sentence patterns that imply hyponymy relations of noun pairs, we showed that we can separate sentences with salient hyponym-hypernym pairs from counterfactuals in which target and feature concepts do not stand in a taxonomic abstraction relationship.
Our setting shows that the level of abstraction in the counterfactual and the semantic similarity of target and feature concepts give rise to different patterns.

Our approach can only give a limited first explanation of the presence of linguistic abstraction within transformers. Firstly, we restrict ourselves to noun pairs and hyponymy, while abstraction comprises many more types of words, relations, and complex mechanisms like frames or scenarios. Further, our experiments cannot explain how the differences in the attention mechanism arise and what they imply. To shed more light on these questions, further research is required, which should analyze both the mathematical theory of the abstraction mechanism and the statistical properties of the input word embeddings. This would make the mechanisms of conceptual abstraction within the transformer architecture more transparent.


\nocite{*}
\section*{Bibliographical References}\label{sec:reference}
\bibliography{abstraction}
\bibliographystyle{lrec-coling2024-natbib}


\end{document}